\documentclass[final]{cvpr}

\usepackage{times}
\usepackage{epsfig}
\usepackage{graphicx}
\usepackage{amsmath}
\usepackage{amssymb}
\usepackage{mmstyle}

\usepackage{booktabs}
\usepackage{tabulary}
\usepackage[dvipsnames]{xcolor}
\usepackage[caption=false]{subfig}
\usepackage[british,english,american]{babel}
\usepackage{soul}
\usepackage{xspace}\xspace
\usepackage{adjustbox}
\usepackage{array}

\usepackage{dblfloatfix}
\usepackage{transparent}

\usepackage{algorithm}
\usepackage{listings}

\usepackage{etoolbox}
\makeatletter
\AfterEndEnvironment{algorithm}{\let\@algcomment\relax}
\AtEndEnvironment{algorithm}{\kern2pt\hrule\relax\vskip3pt\@algcomment}
\let\@algcomment\relax
\newcommand\algcomment[1]{\def\@algcomment{\footnotesize#1}}
\renewcommand\fs@ruled{\def\@fs@cfont{\bfseries}\let\@fs@capt\floatc@ruled
  \def\@fs@pre{\hrule height.8pt depth0pt \kern2pt}%
  \def\@fs@post{}%
  \def\@fs@mid{\kern2pt\hrule\kern2pt}%
  \let\@fs@iftopcapt\iftrue}
\makeatother

\definecolor{citecolor}{HTML}{0071bc}
\usepackage[pagebackref=false,breaklinks=true,letterpaper=true,colorlinks,citecolor=citecolor,bookmarks=false]{hyperref}

\newcommand{\app}{\raise.17ex\hbox{$\scriptstyle\sim$}}

\newcolumntype{x}[1]{>{\centering\arraybackslash}p{#1pt}}
\newcolumntype{y}[1]{>{\raggedright\arraybackslash}p{#1pt}}
\newcolumntype{z}[1]{>{\raggedleft\arraybackslash}p{#1pt}}
\newlength\savewidth\newcommand\shline{\noalign{\global\savewidth\arrayrulewidth
  \global\arrayrulewidth 1pt}\hline\noalign{\global\arrayrulewidth\savewidth}}
\newcommand{\tablestyle}[2]{\setlength{\tabcolsep}{#1}\renewcommand{\arraystretch}{#2}\centering\footnotesize}

\makeatletter\renewcommand\paragraph{\@startsection{paragraph}{4}{\z@}
  {.5em \@plus1ex \@minus.2ex}{-.5em}{\normalfont\normalsize\bfseries}}\makeatother

\def\x{\times}

\usepackage{textpos}  

\def\cvprPaperID{6397} 
\def\confYear{CVPR 2021}

\begin{document}

\title{Instance Localization for Self-supervised Detection Pretraining}

\author{
  Ceyuan Yang$^\dagger$ ~~~~~ Zhirong Wu$^\star$ ~~~~~ Bolei Zhou$^{\dagger,\ddagger}$ ~~~~~ Stephen Lin$^\star$\\
  $^\dagger$Chinese University of Hong Kong ~~~~~~~~~~~ $^\star$Microsoft Research Asia\\
}

\maketitle

\begin{abstract}
   Prior research on self-supervised learning has led to considerable progress on image classification, but often with degraded transfer performance on object detection.
   The objective of this paper is to advance self-supervised pretrained models specifically for object detection.
   Based on the inherent difference between classification and detection, 
   we propose a new self-supervised pretext task, called instance localization. 
   Image instances are pasted at various locations and scales onto background images.
   The pretext task is to predict the instance category given the composited images as well as the foreground bounding boxes.
   We show that integration of bounding boxes into pretraining promotes better task alignment and architecture alignment for transfer learning. 
   In addition, we propose an augmentation method on the bounding boxes to further enhance the feature alignment.
   As a result, our model becomes weaker at Imagenet semantic classification but stronger at image patch localization, with an overall stronger pretrained model for object detection. 
   Experimental results demonstrate that our approach yields state-of-the-art transfer learning results for object detection on PASCAL VOC and MSCOCO\footnote{Code and models are available at \href{https://github.com/limbo0000/InstanceLoc}{this link.} \\ \indent $\ddagger$ indicates corresponding author.}.
\end{abstract}
\section{Introduction}

The dominant paradigm for training deep networks in computer vision is by pretraining and finetuning~\cite{girshick2014rich,long2015fully}.
Typically, the pretraining is optimized to find a single generic representation that is later transferred to various downstream applications.
For example, supervised pretrained models using image-level labels~\cite{krizhevsky2017imagenet,he2016deep} and self-supervised pretrained models by contrastive learning~\cite{he2019momentum} both transfer remarkably well to a number of tasks, e.g., image classification, object detection, semantic segmentation and human pose estimation.

Despite the popularity of this approach, we question the existence of such generic and universal representations for transfer learning.
Recently, it has been observed that self-supervised representations which improve upon image classification performance may fail to translate the advantage to object detection~\cite{caron2020unsupervised,grill2020bootstrap}. 
Also, it is found that high-level features is not what truly matters in transfer to detection and segmentation~\cite{zhao2020makes}. 
These indicate that current self-supervised models may overfit to the classification task while becoming less effective for other tasks of interest.

\begin{figure}[t]
	\centering
	\includegraphics[width=1.0\linewidth]{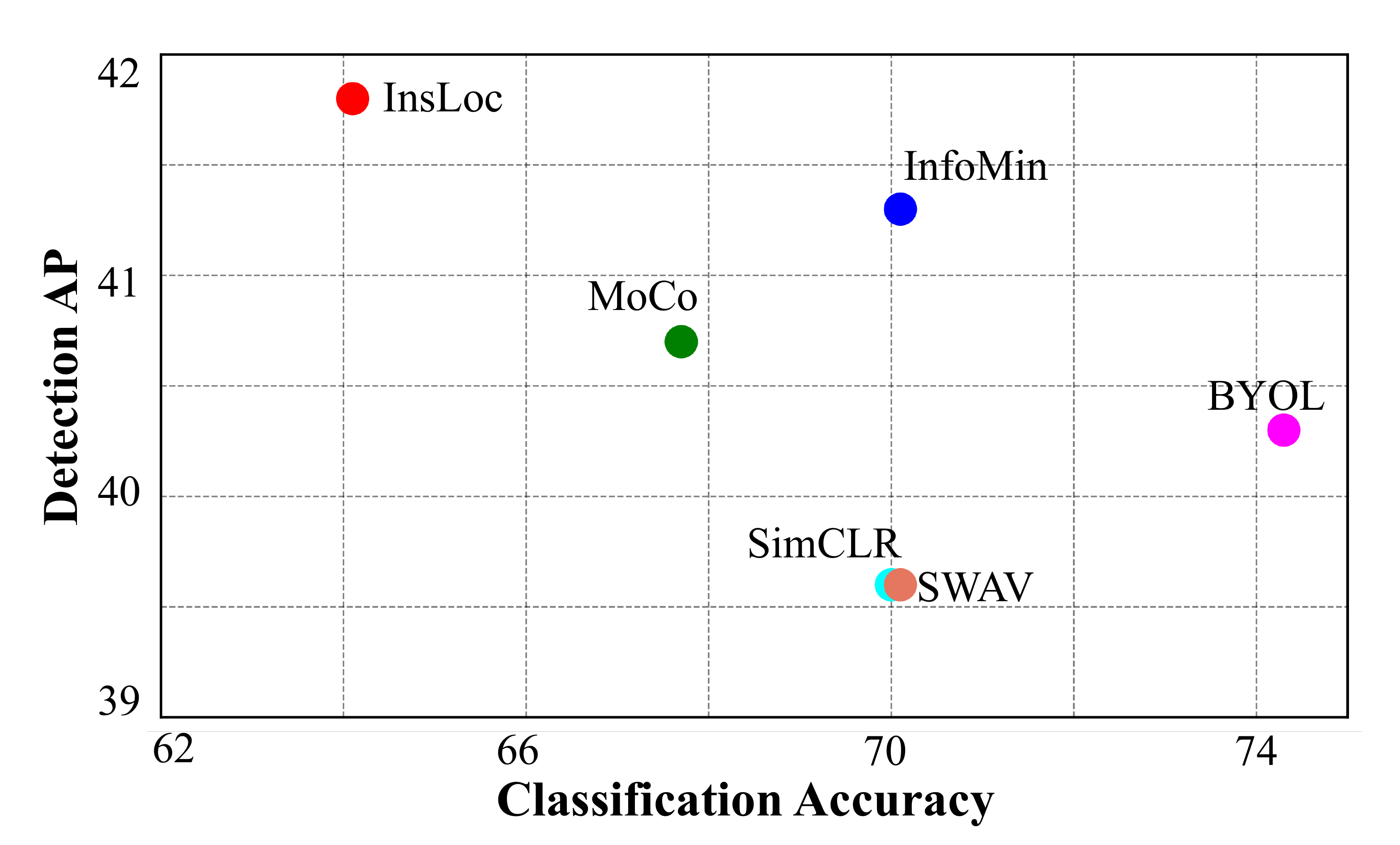}
	\captionsetup{font=small}
	\caption{
	    For visual transfer learning, it is widely assumed that the ImageNet classification accuracy and object detection performance are positively correlated.
	    By studying recent self-supervised models, we find that this is not actually the case.
	    We propose a novel approach, called Instance Localization (InsLoc), which sacrifices ImageNet classification accuracy, but enjoys  better generalization ability for object detection.
	    }
    \label{fig:teaser}
  \end{figure}
We identify two issues that contribute to task misalignment in transfer learning.
The first is that the pretrained network needs to be re-purposed into the target network architecture for finetuning. 
This often involves non-trivial architectural changes, such as inserting a feature pyramid~\cite{lin2017feature} or employing convolution kernels with large dilations~\cite{chen2017deeplab}.
Second, for typical contrastive learning models,
the pretraining pretext task considers an image holistically in instance discrimination~\cite{wu2018unsupervised}, without explicit spatial modelling over regions. 
Though it enhances transferability for classification, this practice is less compatible with spatial reasoning tasks, such as object detection.

In this paper, we propose a new self-supervised pretext task, called instance localization, specifically for the downstream task of object detection.
Akin to instance discrimination, which learns a classifier for individual image instances, 
instance localization additionally takes bounding box information into account for representation learning.
We create our training set by taking crops of foreground images and pasting them at various aspect ratios and scales onto different locations of background images.
Self-supervised pretraining follows by extracting RoI features using bounding boxes and performing contrastive learning using instance labels.
In this way, not only does the network architecture maintain consistency during transfer, but the pretraining task also includes localization modelling, which is crucial for object detection.

Introducing bounding boxes into pretraining encourages explicit alignment between convolutional features and foreground regions.
The feature responses thus become sensitive to translations in the image domain, benefiting detection~\cite{chen2019revisiting}. 
We additionally find that feature alignment can be strengthened by inducing augmentations on the bounding box coordinates.
Specifically, spatially jittered bounding boxes are randomly selected from a set of region proposal anchors.

We implement the approach within the framework of momentum contrast~\cite{he2019momentum}.
The network takes the composited images and bounding boxes as input, and extracts region embeddings for contrastive learning.
Compared with the baseline approach which considers holistic instances,
a linear probe on the last-layer features shows reduced performance for image classification, while achieving improvements in regressing bounding box locations.
Experimentally, we study two popular detection backbone networks, ResNet50-C4 and ResNet50-FPN.
For both backbone networks, our instance localization approach elevates performance substantially, surpassing the state-of-the-art transfer learning results on PASCAL VOC \cite{everingham2010pascal} and MSCOCO \cite{lin2014microsoft}.
Notably, our model is even more advantageous for object detection under the small data regime.

\section{Related Work}
\begin{figure*}[t]
	\centering
	\includegraphics[width=1.0\linewidth]{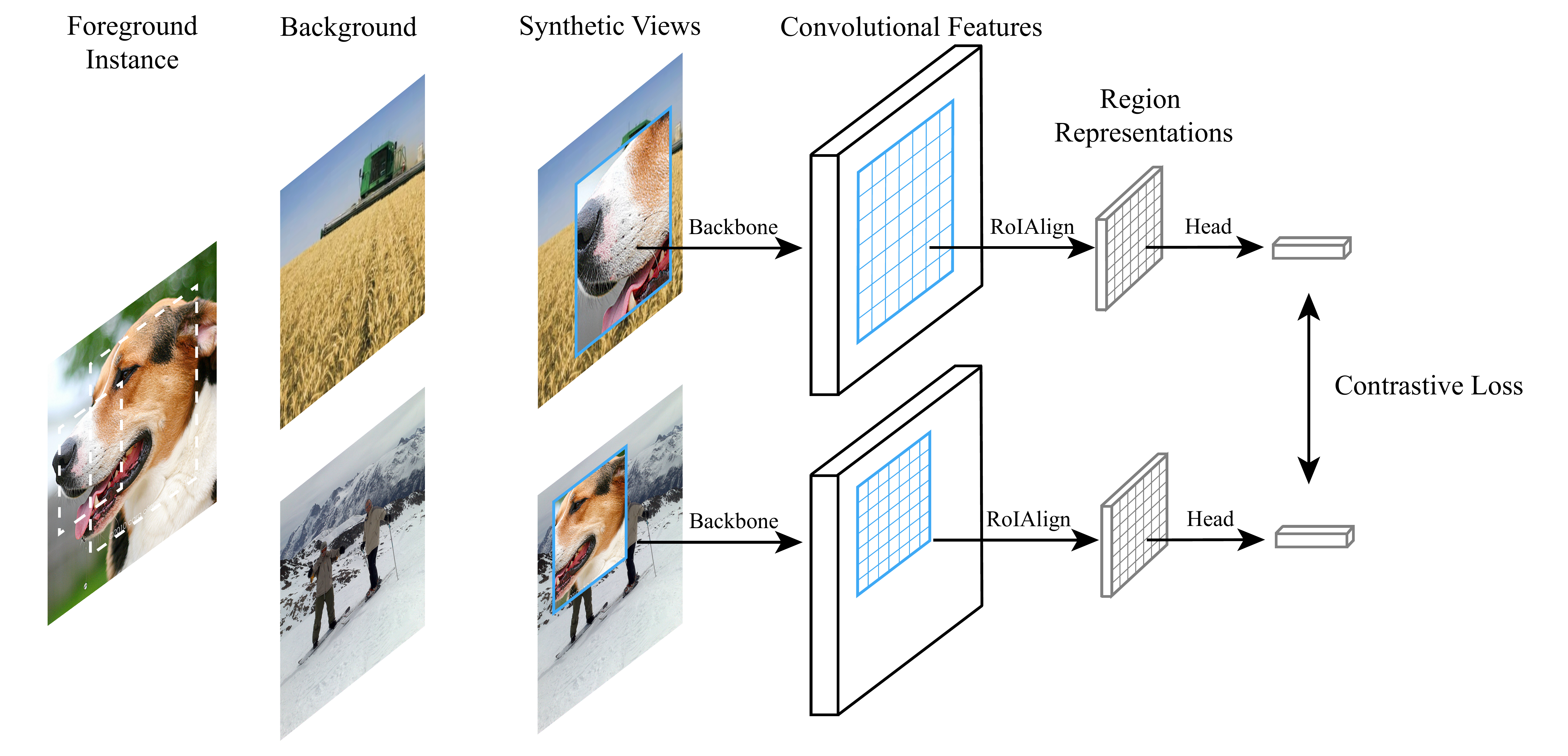}
	\captionsetup{font=small}
	\caption{
	   \textbf{Overview of Instance Localization.}  
	   Given a foreground image instance, we first sample two background images randomly from the image gallery.
	   Two views of the foreground image are generated and copy-pasted onto the corresponding background image.
	   The convolutional network takes each synthetic view, and RoIAlign extracts the region representation using the foreground bounding box coordinates.
	   Contrastive learning follows the region representation.
	   Negative samples are omitted for brevity.
	   }
    \vspace{-1.5em}
       \label{fig:framework}
  \end{figure*}
\textbf{Self-supervised Learning.}
The central idea of self-supervised learning is to create free supervisory labels from visual data, and use the free supervision to obtain generalizable and transferrable representations.
One of the simplest forms of a pretext task is to reconstruct the input image using a generative model.
The latent representation in a generative model is thought to capture the high-level structures and semantic manifolds of the input distribution.
Auto-encoders~\cite{vincent2010stacked} and Boltzmann Machines~\cite{salakhutdinov2009deep} show such capability on handwritten digits, but fail to work on natural images.
Later, the advance of GANs~\cite{zhu2016generative} enabled manipulation of generative content by disentangling neural responses of the latent representation into facial attributes, pose and lighting conditions.
Recent work on BigBiGAN~\cite{donahue2019large} and Image-GPT~\cite{chen2020generative} demonstrate that extremely large generative models may deliver very promising visual recognition representations. However, a fundamental question that remains is how learning to generate image pixels relates to high-level visual understanding.

Aside from reconstructing image pixels, another kind of pretext task is to withhold some part of the data and then predict it from the other part.
Colorization~\cite{zhang2016colorful} withholds color information and attempts to predict it from grayscale values.
Context prediction~\cite{doersch2015unsupervised} splits the spatial content into a 3-by-3 grid of patches. The network is then trained to predict the spatial relationship between patches.
The way the pretext task is formulated strongly affects the knowledge that is learned from the data.
The colorization approach tends to work when objects in the same category share the same color.
Context prediction assumes that objects of one category share the same spatial configurations.
Since different pretext tasks extract visual knowledge of different aspects, a multi-task approach~\cite{doersch2017multi} that combines their individual knowledge boosts the learning performance.

One popular pretext task for self-supervised learning is contrastive learning, or more specifically instance discrimination~\cite{wu2018unsupervised}.
Each instance in the training dataset is treated as a single category of its own.
The learning objective is simply to classify each instance from the rest.
The critical component of contrastive learning is the data augmentation used for inducing invariances~\cite{wu2018unsupervised,he2019momentum,chen2020simple}.
The ideal data augmentations should reflect the intra-class variations,
and commonly-used augmentations include cropping, scaling, color jittering, and blurring. 
Recent research on contrastive learning focuses on developing better augmentations~\cite{caron2020unsupervised}, designing projection head structures~\cite{grill2020bootstrap}, and even alleviating the requirement of negative samples~\cite{grill2020bootstrap}.
While the leading contrastive learning methods BYOL and SwAV push ImageNet performance to an impressive 74\%  with linear readoff classifiers, their transfer performance to object detection actually drops below that of MoCo~\cite{he2019momentum}.
This suggests that these self-supervised methods are overfitted to a single downstream task of classification, sacrificing generalization to other tasks. 

We propose a novel pretext task for self-supervised pretraining, with a focus on transfer to object detection.
Building on top of instance discrimination, we introduce the use of bounding boxes in the pretraining stage.
Towards improved localization, our method learns a representation in which there is alignment between bounding boxes and their corresponding foreground features.
Prior works that explicitly address patch-level spatial modelling include CPC~\cite{oord2018representation} and context prediction~\cite{doersch2015unsupervised}. 
These works reason about spatial arrangement based on patch contents within an image.
In contrast, our pretext task considers spatial relationships between two distinct images composited together.

There is a spectrum of other pretext tasks for self-supervised learning on images and videos,
such as inpainting~\cite{pathak2016context}, rotation prediction~\cite{gidaris2018unsupervised}, jigsaw puzzles~\cite{noroozi2016unsupervised}, and motion segmentation~\cite{pathak2017learning} on images, as well as temporal order~\cite{misra2016shuffle}, temporal speed~\cite{benaim2020speednet}, and synchronization~\cite{arandjelovic2018objects} on videos. 
A detailed survey and description of each pretext task is beyond the scope of the paper.

\textbf{Learning with Image Compositions.}
Creating synthesized imagery by copying a foreground object onto backgrounds is a popular data augmentation technique.
Given the foreground object mask, prior works successfully apply this technique for supervised instance segmentation~\cite{dwibedi2017cut,fang2019instaboost} and unsupervised learning~\cite{zhao2020distilling}.
Our work also synthesizes image compositions, but without a need for object masks or clean contours.

\textbf{Self-Training with Unlabeled Data.}
Besides transfer learning, self-training~\cite{xie2020self,zoph2020rethinking} is a promising direction for utilizing unlabeled data when labelled data is limited.
The idea is to bootstrap the model by using supervised learning on few labelled samples to generate pseudo labels on the unlabeled samples. The model is further optimized with supervised learning jointly on the labels and pseudo labels.
Self-training, however, may become vulnerable when the labelled set is scarce.
Transfer learning and self-training may be integrated, as explored in SimCLR-v2~\cite{chen2020big}.

\section{Pretext Task -- Instance Localization}
  
Image classification favors translation and scale invariance, where
objects of various scales and locations are reduced to a single
discrete variable representing object categories.
In contrast, object detection desires translation and scale equivariance.
Feature representations for object detection should be able to
preserve and reflect information about object sizes and locations.
The inherent difference between the two tasks requires dedicated modeling for each task.
Recent works in contrastive learning focus on designing techniques for image classification. 
Translation and scale invariance are enforced by learning consistency between two random views of an image.
As a result, the pretext task of instance discrimination overfits to holistic classification
and fails to promote spatial reasoning.

We propose a novel pretext task called instance localization (InsLoc) as an extension to instance discrimination.
As illustrated in Figure~\ref{fig:framework},
we synthesize image compositions by overlaying foreground instances onto backgrounds.
The objective is to discriminate the foreground from the background using bounding box information.
In order to achieve this task, one has to localize the foreground instance first and then extract foreground features.

Denote the composited image as $I'$ with the foreground image $I$ overlaid on the bounding box $b$.
The task $\mathcal{T}$ is to predict instance label $y$ for $I$,
\begin{equation}
    y \gets \mathcal{T}(I', b).
\end{equation}

\section{Learning Approach}\label{sec:learning}
  
We aim to learn a representation which is not only semantically powerful, but also equivariant to translation and scale.
We first describe our approach of introducing bounding box representations into a contrastive learning framework in Sec.~\ref{subsec:insloc}.
Data augmentations on the bounding boxes are presented as an effective way to improve localization ability in Sec.~\ref{subsec:box_aug}.
We finally give the architecture details of our approach on two popular detection backbones, R50-C4 and R50-FPN, in Sec.~\ref{subsec:instantiations}.

\begin{figure}[t]
	\centering
	\includegraphics[width=1.0\linewidth]{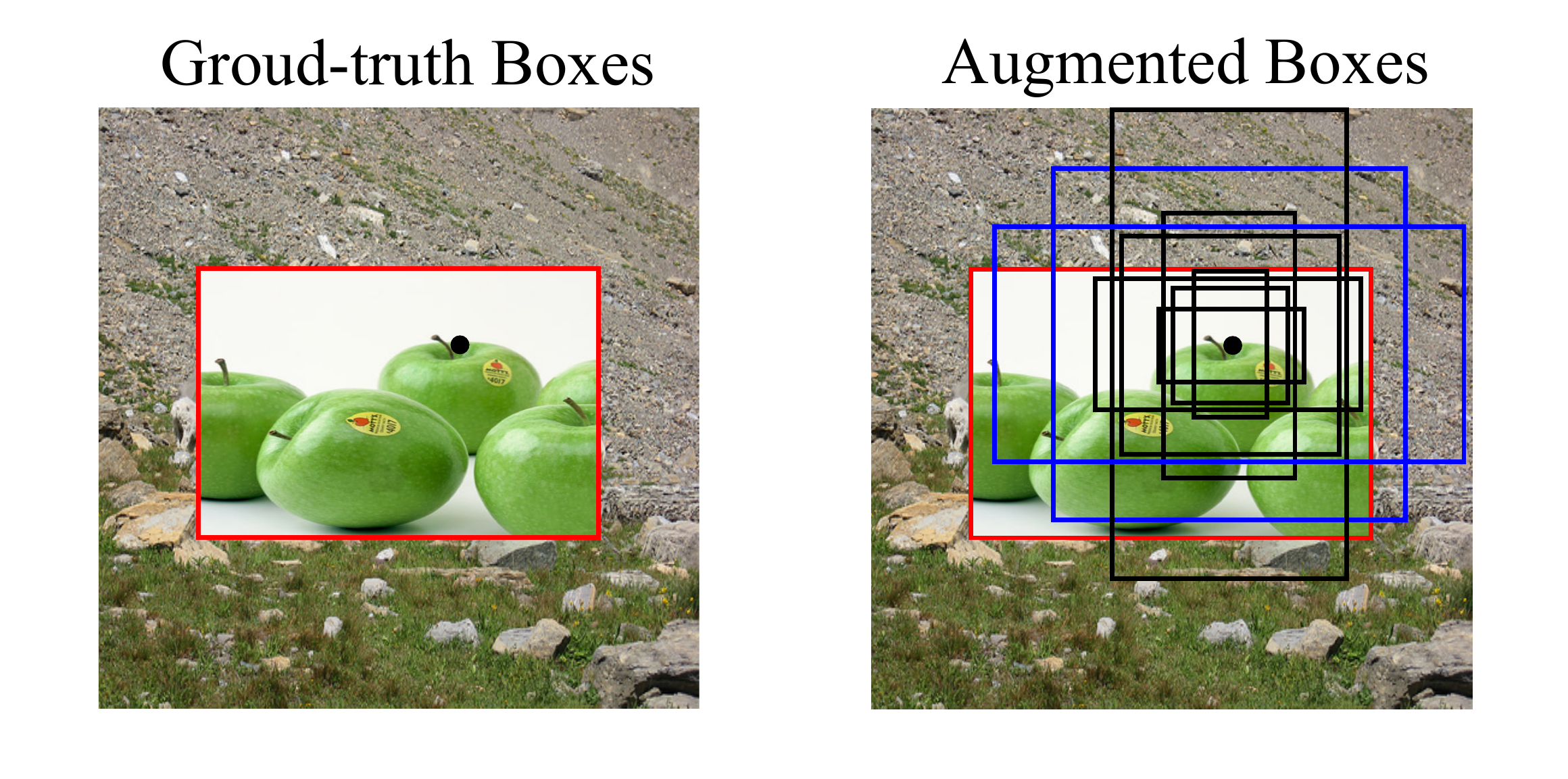}
	\captionsetup{font=small}
	\caption{
	   \textbf{Bounding boxes for spatial modeling.} The \textcolor{red}{red} box denotes the ground-truth bounding box of the foreground image. On the right, we show a set of anchor boxes centered on a single spatial location. By leveraging the multiple anchors with the diverse scales, locations and aspect ratios, we augment the ground truth with the \textcolor{blue}{blue} boxes whose IoU is larger than 0.5.
	   }
       \label{fig:framework}
  \end{figure}

\subsection{Instance Discrimination with Bounding Boxes}\label{subsec:insloc}

\vspace{2pt}
\noindent \textbf{Instance Discrimination.}
Contrastive learning takes two random ``views" as query $I_q$ and key $I_{k_+}$ images, which are derived from random augmentations from the same instance. 
The corresponding features $v_q$ and $v_{k_+}$ are first extracted by a backbone network $f$ (\eg $v_q = f(I_q)$) and then projected to a unit sphere via a head network $\phi$. The contrastive loss, \ie InfoNCE \cite{oord2018representation}, is computed as
\begin{equation}\label{eqa:cl}
    \mathcal{L}  = -\log \frac{\exp(\phi(v_q){\cdot}\phi(v_{k_+}) / \tau)}{\sum_{i=0}^{N}\exp({\phi(v_q)\cdot}\phi(v_{k_i}) / \tau)},
\end{equation}
where $\tau$ and $N$ are the temperature and the number of negative samples, respectively.

\vspace{2pt}
\noindent \textbf{Spatial Modeling with Bounding Boxes.}
We aim to enforce spatial alignment between input regions and convolutional features, along with contrastive learning of discriminative instances.
To do so, given the image $I$, we first sample a random background image $B$, which is simply taken as any other image in the training set.
We then define the composition operation $C$, which copies and pastes a random crop of the image $I$ onto the background $B$ at a random position and scale.
The operation returns the composited image $I'$ and the bounding box parameters $b$,
\begin{align}
     I^{\prime}_{q}, b_{q}     & = C(I_q, B_q), \\
     I^{\prime}_{k_+}, b_{k_+} & = C(I_{k_+}, B_{k_+}),
\end{align}
where $I_q$ and $I_{k_+}$ are crops from the same image instance, and $B_q$ and $B_{k_+}$ are their respective background images\footnote{It should be noted that different background images need to be used for the two views of a foreground instance. Otherwise, the model may cheat on background cues for contrastive learning.}.
In practice, the foreground image is resized with a random aspect ratio\footnote{Empirically, we find $\left[1/3, 3/1\right]$ and $\left[1/2, 2/1\right]$ lead to better performance for R50-C4 and R50-FPN respectively.} and a random scale between 128 to 256 pixels.
With the bounding box parameters $b$, RoIAlign \cite{he2017mask} is applied to extract the foreground features on the convolutional feature maps,
\begin{align}\label{eqa:feat_extract}
     v^{\prime}_{q}  & = \text{RoIAlign}(f(I^{\prime}_q), b_q), \\
     v^{\prime}_{k_{+}}  & = \text{RoIAlign}(f(I^{\prime}_{k_{+}}), b_{k_+}).
\end{align}
With the query and key features, contrastive learning follows similarly to Eq.~\ref{eqa:cl}. Figure~\ref{fig:framework} illustrates our framework.

A problem that complicates detection is the discrepancy between an image region and its spatially corresponding deep features. Since the receptive field of pooled deep features typically extends in the image well beyond the pooling area, the pooled features are influenced by image content outside its vicinity. Consequently, for a bounding box that covers the foreground, its features are affected by the surrounding background, making it harder to localize.

Our instance discrimination with bounding boxes addresses this problem in a data-driven manner. By encouraging similarity between pooled foreground features of the same instance but with different backgrounds, the effective receptive field is learned to match the spatial extent of the bounding box. Establishing this explicit correspondence between convolutional features and their effective receptive field facilitates localization with the learned representation.

\subsection{Bounding-Box Augmentation}\label{subsec:box_aug}

Image augmentations play a key role in contrastive learning of representations~\cite{dosovitskiy2015discriminative,chen2020simple}.
We hypothesize that a similar augmentation strategy may also be effective on the bounding boxes. 
Specifically, jittered boxes around the ground truth location may include regions on the background.
Therefore, representations may be further guided to spatially disregard the background and acquire the localization ability.

\vspace{2pt}
\noindent \textbf{Augmentations as predefined anchors.} 
Instead of directly shifting the bounding box spatially, we leverage the anchors in the region proposal network (RPN) \cite{ren2015faster} to cover the diversity of the augmented boxes. The anchors are a set of pre-defined bounding box proposals with diverse scales, locations and aspect ratios. 
Given a ground truth box, we calculate its IoU against all the anchors.
Anchors with a high overlap (larger than 0.5) are filtered, and a random one is selected as the augmentated box.
Due to the anchor-based design, we are able to obtain a diverse collection of box proposals with a dynamic range of IoUs.
We apply the bounding box augmentations on RoIAlign module of the query encoder while the momentum encoder always uses the ground truth one for pooling.

\subsection{Architectural Alignment}\label{subsec:instantiations}
One critical issue that contributes to the task misalignment in transfer learning is the non-trivial architectural adjustment.
The pretrained network needs to be re-purposed into a detection network by appending region-wise operations and head networks.
Our introduction of bounding box representation allows to minimize the architectural discrepancy between pretraining and finetuning.
To be specific, the RoIAlign operation in pretraining introduces region-wise representations, which tightly mimics the detection behavior in finetuning.
We provide details for the detection architecture R50-C4 and R50-FPN during pretraining.

\vspace{2pt}
\noindent \textbf{R50-C4.}
On the standard ResNet50 architecture, we insert the RoI operation on the output of the 4-th residual block. 
Bounding box coordinates are then used to extract region features.
The entire 5-th residual block is treated as the head network for classifying regions.

\vspace{2pt}
\noindent \textbf{R50-FPN.}
R50-FPN uses lateral connections to form a 4-level feature hierarchy on top of ResNet50. Each level of feature is responsible for modeling objects in a corresponding scale.
We insert the RoI operations on all levels in the FPN hierarchy. 
The instance localization task is performed on all 4 feature levels concurrently~\cite{yang2020vthcl}, where each level maintains a separate memory queue of negative examples in order to avoid cross-level cheating.
In this way, not only the ResNet50 network, but also the FPN layers can be pretrained.

\begin{table}[t]
\setlength{\tabcolsep}{2pt}
\normalsize
\begin{tabular}{x{56}|x{40}|x{40}x{40}x{40}c}
\normalsize
Methods    & Epoch &   AP           & AP$_\text{50}$  & AP$_\text{75}$  \\ 
\shline
Random init  & -     &   33.8         & 60.2            & 33.1 \\
Supervised   & 90     &   53.5         & 81.3            & 58.8 \\
\hline
Relative Loc.& 200   &   50.6         & 76.9            & 55.2 \\
MoCo-v2      & 200   &   57.0         & 82.4            & 63.6 \\
MoCo-v2      & 800   &   57.4         & 82.5            & 64.0 \\
InfoMin      & 200   &   57.6         & 82.7            & 64.6 \\
InfoMin      & 800   &   57.5         & 82.7            & 64.3 \\
SimCLR       & 200   &   51.5         & 79.4            & 55.6 \\
BYOL         & 300   &   51.9         & 81.0            & 56.5 \\
SwAV         & 400   &   45.1         & 77.4            & 46.5 \\
\hline
\textbf{InsLoc}   & 200   & 57.9   & 82.9   & 64.9 \\
\textbf{InsLoc}   & 400   &\textbf{58.4}   & \textbf{83.0}   & \textbf{65.3}
\end{tabular}
\vspace{1pt}
\caption{
    \normalsize
    \textbf{Object detection on PASCAL VOC}.
    Models are fine-tuned on \texttt{trainval07+12} and tested on \texttt{test2007}. 
    We evaluate the SimCLR, BYOL, SwAV models ourselves, while reporting the remaining results from their original papers.
    All numbers are averaged over five trials.
}
\label{tab:exp_voc}
\end{table}
\begin{table*}[t]
\hspace{-1.2em}
\resizebox{1.02\linewidth}{!}{
\subfloat[Mask R-CNN, R50-\textbf{C4}, \textbf{2$\x$} schedule]{
\tablestyle{.8pt}{1.05}
\label{tab:sota_2x:c4}
\begin{tabular}{x{40}|x{25}|x{25}x{25}x{25}|x{25}x{25}x{25}c}
Methods    & Epoch &  AP$^{bb}$  & AP$_\text{50}^{bb}$  & AP$_\text{75}^{bb}$ & AP$^{mk}$  & AP$_\text{50}^{mk}$  & AP$_\text{75}^{mk}$   \\ 
\shline
Random                & -     &   35.6         & 54.6            & 38.2         &   31.4          & 51.5            & 33.5\\
Supervised            & 90     &   40.0         & 59.9            & 43.1         &   34.7          & 56.5            & 36.9 \\
\hline
Rel. Loc.             & 200   &   38.0         & 57.4            & 41.0         &   33.3          & 54.1            & 35.4 \\
MoCo-v2               & 200   &   40.7         & 60.5            & 44.1         &   35.6          & 57.4            & 37.1 \\
MoCo-v2               & 800   &   41.2         & 60.9            & 44.6         &   35.8          & 57.7            & 38.2 \\
InfoMin               & 200   &   41.3         & 61.2            & 45.0         &   36.0          & 57.9            & 38.3 \\
InfoMin               & 800   &   41.2         & 61.2            & 44.8         &   35.9          & 57.9            & 38.4 \\
SimCLR                & 200   &   39.6         & 59.1            & 42.9         &   34.6          & 55.9            & 37.1 \\
BYOL                  & 300   &   40.3         & 60.5            & 43.9         &   35.1          & 56.8            & 37.3 \\
SWAV                  & 400   &   39.6         & 60.1            & 42.9         &   34.7          & 56.6            & 36.6 \\
\hline
\textbf{InsLoc}       & 200   &   41.4         & 60.9            & 45.0         &   35.9          & 57.6            & 38.4 \\
\textbf{InsLoc}       & 400   & \textbf{41.8}   & \textbf{61.6}   & \textbf{45.4} &\textbf{36.3}   & \textbf{58.2}   & \textbf{38.8} 
\end{tabular}
}  

\subfloat[Mask R-CNN, R50-\textbf{FPN}, \textbf{2$\x$} schedule]{
\tablestyle{.8pt}{1.05}
\label{tab:sota_2x:fpn}
\begin{tabular}{x{25}x{25}x{25}|x{25}x{25}x{25}c}
  AP$^{bb}$  & AP$_\text{50}^{bb}$  & AP$_\text{75}^{bb}$ & AP$^{mk}$  & AP$_\text{50}^{mk}$  & AP$_\text{75}^{mk}$   \\ 
\shline
   38.4         & 57.5            & 42.0            &   34.7         & 54.8            & 37.2 \\
   41.6         & 61.7            & 45.3            &   37.6         & 58.7            & 40.4 \\
\hline
   39.4         & 58.7            & 42.7            &   35.6         & 55.9            & 38.1 \\
   41.7         & 61.6            & 45.6            &   37.6         & 58.7            & 40.5 \\
   42.5         & 62.3            & 46.8            &   38.2         & 59.6            & 41.1 \\
   42.5         & 62.7            & 46.8            &   38.4         & 59.7            & 41.4 \\
   42.1         & 62.3            & 46.2            &   38.0         & 59.5            & 40.8 \\
   40.8         & 60.6            & 44.4            &   36.9         & 57.8            & 39.8 \\
   42.3         & 62.6            & 46.2            &   38.3         & 59.6            & 41.1 \\
   42.3         & 62.8            & 46.3            &   38.2         & 60.0            & 41.0 \\
\hline
   43.2         & 63.5           & \textbf{47.5}    &   38.7         & 60.5            & \textbf{41.9} \\
\textbf{43.3}   & \textbf{63.6}  & 47.3   & \textbf{38.8}            & \textbf{60.9}    & 41.7
\end{tabular}	
}  
}  

\vspace{.3em}
\caption{\textbf{Object detection and instance segmentation on COCO.} Models are fine-tuned on \texttt{train2017} and tested on \texttt{val2017}.
}
\label{tab:sota_2x}
\end{table*}

\section{Experimental Results}\label{sec:exp}

We evaluate the generalization ability of our model for transfer learning on the mainstream object detection benchmarks: PASCAL VOC~\cite{everingham2010pascal} and MSCOCO~\cite{lin2014microsoft}. 
The main experimental results with state-of-the-art comparisons are presented in Section~\ref{subsec:results}. 
Ablation studies and discussions regarding the trade-offs between semantic classification and localization are conducted in Section~\ref{subsec:ablation}. 
In Section~\ref{subsec:minicoco}, we present an experiment on the mini version of MSCOCO to demonstrate the fast generalization ability of our model under a small amount of labelled data.

\noindent \textbf{Dataset.} 
The ImageNet dataset~\cite{deng2009imagenet} with 1.3 million images is used for pretraining, while
PASCAL VOC~\cite{everingham2010pascal} and MSCOCO~\cite{lin2014microsoft} are used for transfer learning.
PASCAL VOC0712 contains about 16.5K images with bounding box annotations in 20 object categories.
MSCOCO contains about 118K images with bounding box and instance segmentation annotations in 80 object categories.

\noindent \textbf{Pretraining.} 
We largely follow the hyper-parameters from the official implementation of MoCo-v2~\cite{chen2020improved}.
We optimize the model with synchronized SGD over 8 GPUs with a weight decay of 0.0001, a momentum of 0.9, and a batch size of 32 on each GPU. 
The optimization takes 200-400 epochs with an initial learning rate of 0.03 and a cosine learning rate schedule~\cite{loshchilov2016sgdr}. 
A two-layer MLP head is used for contrastive learning, and the temperature parameter is set to $0.2$ in Eq~\ref{eqa:cl}. 
We also maintain a memory queue of 65536 negative samples. 
The momentum coefficient is set to 0.999 for updating the key encoder.

\noindent \textbf{Data augmentations.}
During pretraining, image augmentations for the foreground content follow MoCo-v2~\cite{chen2020improved}. 
Specifically, we apply random resized cropping, color jittering, grayscaling, gaussian blurring and horizontal flipping. 
Even stronger augmentations may further boost the transfer performance~\cite{tian2020makes, chen2020simple} but lie outside the focus of our work.  

\noindent \textbf{Fine-tuning.} 
The backbone network is transferred from the pretraining task to the downstream task.
Following MoCo-v2 \cite{chen2020improved}, synchronized batch normalization is used across all layers including the newly initialized batch normalization layers. Detectors are implemented and fine-tuned using detectron2~\cite{wu2019detectron2}.

\subsection{Main Results}\label{subsec:results}
We provide our experimental results for object detection and compare the performance with state-of-the-art methods.
The pretrained weights of SimCLR~\cite{chen2020simple} and BYOL~\cite{grill2020bootstrap} are borrowed from a third-party implementation\footnote{ \url{https://github.com/open-mmlab/OpenSelfSup/blob/master/docs/MODEL_ZOO.md}}, while those of MoCo~\cite{he2019momentum}, InfoMin~\cite{tian2020makes} and SwAV~\cite{caron2020unsupervised} are collected from their official implementations.

\subsubsection{PASCAL VOC Object Detection}
\noindent \textbf{Setup.}
We use the Faster R-CNN detector~\cite{ren2015faster} with a R50-C4 backbone architecture. 
Optimization takes a total of 24k iterations.
The learning rate is initialized to 0.02 and decayed to be 10 times smaller after 18k and 22k iterations.
The image scale is within [480, 800] pixels for training and set to 800 at inference.  
AP, AP$_\text{50}$ and AP$_\text{75}$ are shown as the evaluation metrics.

\noindent \textbf{Results.}
The transfer results are summarized in Table~\ref{tab:exp_voc}. 
All values are averaged over five independent trials due to large variance. 
We report our results under 200 epochs and 400 epochs of pretraining.
Compared with our direct baseline, MoCo-v2~\cite{chen2020improved}, our model obtains an improvement of \textbf{+0.9} and \textbf{+1.0} AP with 200 and 800 epochs, respectively.
It also outperforms all previous approaches without using complex and stronger data augmentations such as RandAugment or Multi-crop.
Our pretrained model obtains the state-of-the-art results on this benchmark.

\subsubsection{COCO Object Detection and Segmentation}
\noindent \textbf{Setup.}
We use the Mask R-CNN~\cite{he2017mask} framework with R50-C4 and R50-FPN backbone networks. 
Since previous literature~\cite{he2019rethinking} suggests that a detector with random initialization can match the supervised counterpart on COCO~\cite{lin2014microsoft} when the training schedule is very long, we conduct this transfer experiment on 2$\times$ schedules with 180k iterations of optimization.
The learning rate is initialized to 0.02 and decayed to be 10 times smaller after 120k and 160k iterations.
The image scale is within [640, 800] pixels for training and set to 800 for testing. 
AP, AP$_\text{50}$ and AP$_\text{75}$ are shown as the evaluation metrics for bounding box detection and instance segmentation.

\noindent \textbf{Results.}
Table~\ref{tab:sota_2x} shows the results for R50-C4 (Table~\ref{tab:sota_2x:c4}) and R50-FPN (Table~\ref{tab:sota_2x:fpn}). AP$^{bb}$ and AP$^{mk}$ denote the AP of bounding box detection and instance mask segmentation, respectively. 
Pretrained for 200 epochs, InsLoc outperforms the direct baseline  MoCo-v2~\cite{he2019momentum} by \textbf{+0.7} and \textbf{+1.5} AP for the R50-C4 and R50-FPN backbones.
Pretrained for 400 epochs, InsLoc reaches the new state-of-the-art performance, surpassing all prior self-supervised models with possibly stronger image augmentations. In particular, InsLoc introduces the significant improvements over fully supervised ImageNet pretraining, \ie \textbf{+1.8} and \textbf{+1.7} AP for R50-C4 and R50-FPN respectively.
Notably, InfoMin shows reduced transfer performance when the model is pretrained longer.
BYOL and SwAV are competitive for the R50-FPN backbone but relatively weaker for the R50-C4 backbone.
Our model is consistently stronger in all aspects.

\begin{table}[t]
\small
\resizebox{1.0\linewidth}{!}{
\subfloat[Semantic \vs Localization. The linear readout accuracy of linear classification (Cls) and localization (Loc) as well as overall fine-tuned detection AP are presented. Detector architecture is R50-C4.]
{
\label{tab:ablation:alginment}
\begin{tabular}{x{56}|x{25}x{25}|x{25}x{25}c}
Methods           &  Cls &   Loc  & AP$^{bb}$ & AP$^{mk}$ \\ 
\shline
SWAV             & 70.1       &   58.4       & 34.0 & 30.4     \\
BYOL             & 74.3       &   67.6       & 37.5 & 32.8     \\
MoCo-v2             & 67.7       &   71.9       & 38.9      & 34.1     \\
\hline
InsLoc              & 61.7       &   74.2       & 39.5      & 34.5 
\end{tabular}	
}  
}  
\small
\resizebox{1.0\linewidth}{!}{
\subfloat[RoiAlign (RA) is inserted into the baseline to reflect architectural changes. The instance localization task is then performed, \ie foreground images are copied and pasted (CP) onto the background images to learn the spatial alignment. Bounding-box augmentation (BBA) is finally applied. The experiments are performed on the R50-FPN architecture.]{
\label{tab:ablation:box_aug}
\begin{tabular}{x{22}x{22}x{22}|x{45}|x{45}c}
RA & CP & BBA & AP$^{bb}$ & AP$^{mk}$ \\ 
\shline
     &    &     &  39.8     &  36.1      \\
\hline
 \checkmark    &             &            &  40.2     &  36.4       \\
 \checkmark    & \checkmark  &            &  41.1     &  36.9      \\
 \checkmark    & \checkmark  & \checkmark &  41.4     &  37.1      \\
\end{tabular}
}  
}  

\vspace{.3em}
\caption{\textbf{Ablation Studies.} 
All numbers are reported with 1$\times$ schedule on the COCO \texttt{val2017} set.}
\label{tab:ablation}
\vspace{-.2em}
\end{table}
\begin{table*}[t]
\normalsize
\resizebox{1.0\linewidth}{!}{
\subfloat[Mask R-CNN, R50-\textbf{C4}, \textbf{1$\x$} schedule]{
\tablestyle{.8pt}{1.05}
\begin{tabular}{x{40}|x{25}|x{25}x{25}x{25}|x{25}x{25}x{25}c}
Methods    & Epoch &  AP$^{bb}$  & AP$_\text{50}^{bb}$  & AP$_\text{75}^{bb}$ & AP$^{mk}$  & AP$_\text{50}^{mk}$  & AP$_\text{75}^{mk}$   \\ 
\shline
MoCo-v2               & 200   &   38.9         & 58.6            & 41.9         &   34.1          & 55.5            & 36.0 \\
MoCo-v2               & 800   &   39.3         & 58.9            & 42.5         &   34.3          & 55.7            & 36.5 \\
\hline
\textbf{InsLoc}       & 200   &   39.5         & 59.1            & 42.7         &   34.5          & 56.0            & 36.8 \\
\textbf{InsLoc}       & 400   &\textbf{39.8}   & \textbf{59.6}   & \textbf{42.9} &\textbf{34.7}   & \textbf{56.3}   & \textbf{36.9} 
\end{tabular}	
}  

\subfloat[Mask R-CNN, R50-\textbf{C4}, \textbf{2$\x$} schedule]{
\tablestyle{.8pt}{1.05}
\begin{tabular}{x{25}x{25}x{25}|x{25}x{25}x{25}c}
  AP$^{bb}$  & AP$_\text{50}^{bb}$  & AP$_\text{75}^{bb}$ & AP$^{mk}$  & AP$_\text{50}^{mk}$  & AP$_\text{75}^{mk}$   \\ 
\shline
   40.7         & 60.5            & 44.1         &   35.6          & 57.4            & 37.1 \\
   41.2         & 60.9            & 44.6         &   35.8          & 57.7            & 38.2 \\
\hline
   41.4         & 60.9            & 45.0         &   35.9          & 57.6            & 38.4 \\
 \textbf{41.8}   & \textbf{61.6}   & \textbf{45.4} &\textbf{36.3}   & \textbf{58.2}   & \textbf{38.8}
\end{tabular}	
}  
}  
\resizebox{1.0\linewidth}{!}{
\subfloat[Mask R-CNN, R50-\textbf{FPN}, \textbf{1$\x$} schedule]{
\tablestyle{.8pt}{1.05}
\begin{tabular}{x{40}|x{25}|x{25}x{25}x{25}|x{25}x{25}x{25}c}
Methods    & Epoch &  AP$^{bb}$  & AP$_\text{50}^{bb}$  & AP$_\text{75}^{bb}$ & AP$^{mk}$  & AP$_\text{50}^{mk}$  & AP$_\text{75}^{mk}$   \\ 
\shline
MoCo-v2               & 200   &   39.8         & 59.4            & 43.6         &   36.1          & 56.5            & 38.9 \\
MoCo-v2               & 800   &   40.4         & 60.2            & 44.2         &   36.4          & 57.2            & 38.9 \\
\hline
\textbf{InsLoc}       & 200   &   41.4         & 61.7            & 45.0         &   37.1          & 58.5            & 39.6  \\
\textbf{InsLoc}       & 400   &   \textbf{42.0}& \textbf{62.3}   & \textbf{45.8}&   \textbf{37.6} & \textbf{59.0}   & \textbf{40.5}  
\end{tabular}	
}  

\subfloat[Mask R-CNN, R50-\textbf{FPN}, \textbf{2$\x$} schedule]{
\tablestyle{.8pt}{1.05}
\begin{tabular}{x{25}x{25}x{25}|x{25}x{25}x{25}c}
  AP$^{bb}$  & AP$_\text{50}^{bb}$  & AP$_\text{75}^{bb}$ & AP$^{mk}$  & AP$_\text{50}^{mk}$  & AP$_\text{75}^{mk}$   \\ 
\shline
   41.7         & 61.6            & 45.6            &   37.6         & 58.7            & 40.5 \\
   42.5         & 62.3            & 46.8            &   38.2         & 59.6            & 41.1 \\
\hline
   43.2         & 63.5           & \textbf{47.5}    &   38.7         & 60.5            & \textbf{41.9} \\
\textbf{43.3}   & \textbf{63.6}  & 47.3   & \textbf{38.8}            & \textbf{60.9}    & 41.7 
\end{tabular}	
}  
}  

\vspace{.3em}
\caption{
\textbf{Baseline comparison with MoCo-v2 for object detection and instance segmentation on COCO}. R50-C4 and R50-FPN backbones are fine-tuned under 1$\times$ and 2$\times$ schedule.
}
\label{tab:longer}
\end{table*}

\subsection{Ablation Study}\label{subsec:ablation}
To further understand the advantages of instance localization,
we conduct a series of ablation studies that examine semantic and localization trade-offs, the effects of the new pretext task, fine-tuning with longer schedules, and pretraining under more optimization epochs.
  
\vspace{2pt}

\noindent \textbf{Whether the improvement is due to stronger semantic features.}
Recent methods tend to focus on linear object classification as the core evaluation metric for learned representations, based on an assumption that representations with stronger semantics always translate well to other downstream tasks.
To further investigate and understand the improvement of the proposed method for object detection, we devise a new readout task for evaluating the localization ability of pretrained models.

\begin{figure}[t]
	\centering
	\includegraphics[width=1.0\linewidth]{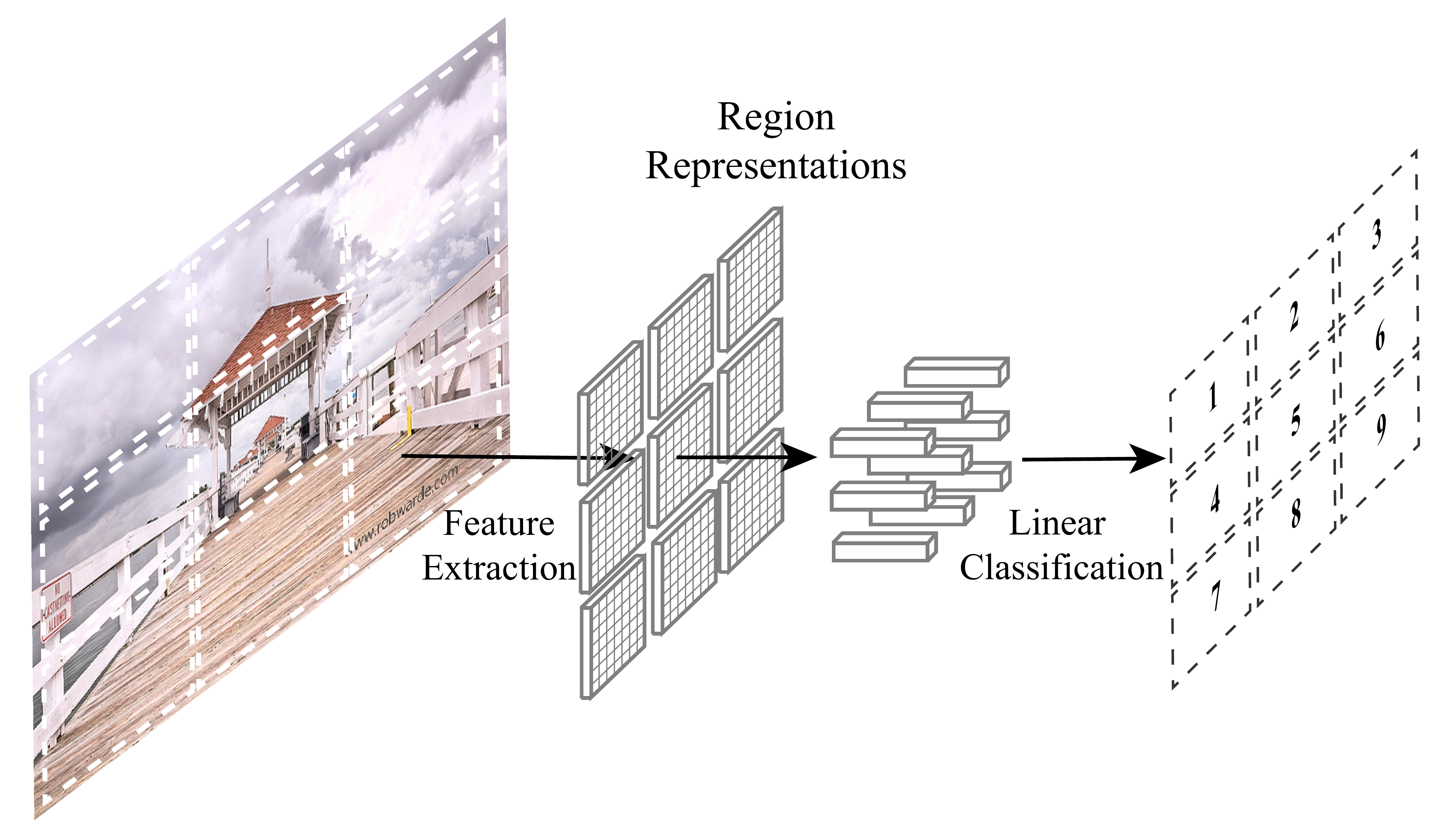}
	\captionsetup{font=small}
	\caption{
	   \textbf{Linear localization evaluation.} 
	   We split a natural image into a grid of region patches.
	   For each region, we extract its vector representation and train a linear classifier to predict the region index in the full image.
	   }
    \label{fig:box_aug}
  \end{figure}

Specifically, given an input image, we split the whole image into $M$ patches. 
The task is to predict the location of each patch based on the region features of the patch using linear classifiers. 
Figure~\ref{fig:box_aug} illustrates this task for $M$ equal to 9.
While earlier work on context prediction~\cite{doersch2015unsupervised} predicts the relative spatial position between two patches,
our evaluation task instead considers the spatial arrangement of a patch with respect to the full image.
For each patch, we extract its vector representation by forwarding it through the backbone network, extracting RoI features, and passing them through the head network.
We append a linear classifier to predict the patch index.
We argue that this task is akin to the detection pipeline, and reflects the localization ability of pretrained models.

Table~\ref{tab:ablation:alginment} shows a comparison of semantic and localization accuracy. 
Instance Localization introduces a clear improvement of $2.3\%$ for the linear localization task, while under-performing MoCo-v2 in linear classification by $6.0\%$. 
This suggests that the overall improvement on object detection is mainly brought by better spatial localization, instead of stronger semantics. 
These results also match a recent finding~\cite{zhao2020makes} that self-supervised pretraining does not transfer high-level semantics for object detection,
but rather low-level and mid-level transfer matters more.
We further include entries for BYOL and SWAV on this localization evaluation in Table~\ref{tab:ablation:alginment}.
Their poor localization ability limits the effectiveness of transfer to object detection.

\vspace{2pt}
\noindent \textbf{Effectiveness of instance localization pretext task.}
Table~\ref{tab:ablation:box_aug} presents the ablation studies of multiple components: the architectural alignment, the instance localization task and the proposed bounding-box augmentation. 
We first integrate the RoiAlign operator to alleviate the architectural changes for the baseline model. To be specific, holistic representations are pooled and extracted from the network, and contrastive learning follows. Also, multiple contrastive losses on the FPN hierarchy (Sec.\ref{subsec:instantiations}) are applied, leading to an overall $+0.4$ AP$^{bb}$ improvement. Such improvements demonstrate the effectiveness to align the architecture for the detection transfer.
We then apply the instance localization task on the composited images with copy-paste operations and bounding box representations, the performance reaches to 41.1 AP$^{bb}$, showing a clear margin over MoCo-v2 by \textbf{+1.3}. 
Finally, when spatial jittering is applied on the bounding boxes, the result is further boosted to $41.4$ AP$^{bb}$. These results strongly verify the effectiveness of the new pretext task \ie instance localization and the corresponding augmentation.

\vspace{2pt}
\noindent \textbf{Effects of fine-tuning schedules.}
Fine-tuning the downstream object detection task with increasing number of iterations may improve the object detection performance.
We examine how the fine-tuning schedule affects the relative improvement of pretrained models.
In Table~\ref{tab:longer}, we study the object detection transfer to COCO under the 1x and 2x fine-tuning schedules.
Using R50-C4, an improvement of 0.6 AP$^{bb}$ with the 1$\times$ schedule translates to an improvement of 0.7 AP$^{bb}$ with the 2$\times$ schedule.
Similar observations were obtained for R50-FPN.
These results show that longer fine-tuning may not significantly weaken the relative improvements, demonstrating the utility of pretrained models for transfer learning.

\vspace{2pt}
\noindent \textbf{Effects of longer pretraining.}
ImageNet linear classification accuracy benefits substantially from longer pretraining. For example, MoCo-v2 improves from $67.5\%$ to $71.1\%$ by increasing the number of pretraining epochs from 200 to 800.
However, for object detection,
longer pretraining may be harmful as shown for InfoMin~\cite{tian2020makes}.
In Table~\ref{tab:longer}, we report the transfer performance on COCO with pretraining models for 400 epochs of optimization.
Compared to the models pretrained for 200 epochs, longer pretraining obtains a consistent improvement and a new state-of-the-art performance.
Even longer pretraining with 800 epochs is computationally expensive and we leave it for future work.

\subsection{Evaluation on Mini COCO}
\label{subsec:minicoco}
Transfer learning to COCO may be of limited significance due to the scale of its dataset. 
Previous literature~\cite{chen2017rethinking} also suggests that training from scratch on COCO with very long learning schedule could provide a strong baseline. 
To demonstrate the generalization ability of our pretrained model under a small amount of labeled data, we conduct an experiment on a mini version of the COCO dataset.

\vspace{2pt}
\noindent \textbf{Dataset.} We randomly select 10\% of the training data (around 11.8K images) from the original \texttt{train2017} set as Mini COCO. The total training data is similar to that of PASCAL VOC~\cite{everingham2010pascal}.
The large variances of objects in terms of scales and aspect ratios remains particularly challenging. We use the full validation set (\ie \texttt{val2017}) of MSCOCO \cite{lin2014microsoft} which contains 5K annotated images for evaluation. 

\vspace{2pt}
\noindent \textbf{Fine-tuning.} Fine-tuning protocols remain the same as for the entire COCO. We use the R50-C4 backbone and finetune the network for 12 epochs. An additional batch normalization layer is inserted after the last residual block.

\vspace{2pt}
\noindent \textbf{Results.}
Table~\ref{tab:mini_coco} summarizes the results. 
We obtain a large improvement of 3.3 AP$^{bb}$ and 2.4 AP$^{mk}$ against MoCo-v2, and 
3.1 AP$^{bb}$ and 2.3 AP$^{mk}$ against the supervised method, demonstrating superior generalization and transfer ability.
Note that the gain on Mini COCO is much greater than the gain on the original COCO.
Such results clearly show that our pretrained model is more data-efficient for transfer learning.

\begin{table}[t]
\normalsize
\begin{tabular}{x{56}|x{40}|x{40}x{40}c}
Methods           &  Epoch    & AP$^{bb}$    & AP$^{mk}$ \\ 
\shline
Supervised           &  90        & 22.9         & 21.2     \\
\hline
Relative Loc.        &  200      & 17.2         & 16.1     \\
MoCo-v2              &  200      & 22.7         & 21.1     \\
InfoMin              &  200      & 23.6         & 21.7     \\
SimCR                &  200      & 20.0         & 18.9     \\
BYOL                 &  300      & 20.6            & 19.6     \\
SWAV                 &  400      & 14.9         & 15.2     \\
\hline
\textbf{InsLoc}      &  200      &\textbf{26.0} &\textbf{23.5}   
\end{tabular}
\vspace{.3em}
\caption{
    \normalsize
    \textbf{Object detection on Mini COCO.} Models are fine-tuned on $10\%$ of COCO \texttt{train2017} for 12 epochs and evaluated on \texttt{val2017}. 
}
\label{tab:mini_coco}
\end{table}
\section{Conclusion}

We propose a new pretext task of instance localization, and introduce the use of bounding boxes in self-supervised representation learning.
The pretrained model is shown to perform weaker for holistic image classification, but much stronger for patch localization.
When transferred to object detection, it achieves notable improvements against the baseline MoCo and obtains the state-of-the-art results on VOC and COCO.
We also show that our approach obtains a larger gain when the labeled data is particularly small.
The experimental results demonstrate that 
transfer performance for object detection can be strengthened
by improving task alignment. 

\vspace{2pt}
\noindent \textbf{Acknowledgments.} 
This project was partially supported by the Centre for Perceptual and Interactive Intelligence (CPII) Ltd under the Innovation and Technology Fund.
We also thank Nanxuan Zhao, Yinghao Xu and Bo Dai for insightful discussions.

{\small
\bibliographystyle{ieee_fullname}
\bibliography{egbib}
}

\end{document}